\documentclass{article}
\usepackage{spconf,amsmath,graphicx}
\usepackage{amsfonts}
\usepackage{booktabs}  
\usepackage{threeparttable}
\usepackage{tablefootnote}
\usepackage{subfigure}
\usepackage{longtable}
\usepackage{xcolor}
\usepackage{hyperref}

\DeclareMathOperator{\tr}{tr}
\DeclareMathOperator{\maximize}{maximize}
\DeclareMathOperator{\minimize}{minimize}

\newtheorem{definition}{Definition}[section]

\title{Deep Deterministic Information Bottleneck \\ with Matrix-based Entropy Functional}
\begin{document}

\name{Xi Yu$^1$, Shujian Yu\sthanks{Contact author: yusj9011@gmail.com.}$^{2}$, Jos\'{e} C. Pr\'{i}ncipe\sthanks{This work was funded in part by the U.S. ONR under grant N00014-18-1-2306 and DARPA under grant FA9453-18-1-0039.}$^1$}
\address{$^1$Computational NeuroEngineering Laboratory, University of Florida, Gainesville, FL 32611, USA\\
$^2$NEC Laboratories Europe, 69115 Heidelberg, Germany}

%

\maketitle
\begin{abstract}
We introduce the matrix-based R{\'e}nyi's $\alpha$-order entropy functional to parameterize Tishby \emph{et al.} information bottleneck (IB) principle~\cite{tishby99information} with a neural network. We term our methodology Deep Deterministic Information Bottleneck (DIB), as it avoids variational inference and distribution assumption. We show that deep neural networks trained with DIB outperform the variational objective counterpart and those that are trained with other forms of regularization, 
in terms of generalization performance and robustness to adversarial attack. Code available at~\url{https://github.com/yuxi120407/DIB}.
\end{abstract}
\section{Introduction}
\label{sec:intro}


The information bottleneck (IB) principle was introduced by Tishby \emph{et al.}~\cite{tishby99information} as an information-theoretic framework for learning. It considers extracting information about a target signal $Y$ through a correlated observable $X$. The extracted information is quantified by a variable $T$, which is (a possibly randomized) function of $X$, thus forming the Markov chain $Y \leftrightarrow X \leftrightarrow T$. Suppose we know the joint distribution $p(X,Y)$, the objective is to learn a representation $T$ that maximizes its predictive power to $Y$ subject to some constraints on the amount of information that it carries about $X$:
\begin{equation}
    \mathcal{L}_{IB}=I(Y;T) - \beta I(X;T),
\end{equation}
where $I(\cdot;\cdot)$ denotes the mutual information. $\beta$ is a Lagrange multiplier that controls the trade-off between the \textbf{sufficiency} (the performance on the task, as quantified by $I(Y;T)$) and the \textbf{minimality} (the complexity of the representation, as measured by $I(X;T)$). In this sense, the IB principle also provides a natural approximation of \emph{minimal sufficient statistic}. 

The IB principle is appealing, since it defines what we mean by a good representation through a fundamental trade-off. However, solving the IB problem for complicated $p(X,Y)$ is often criticized to be hard or impossible~\cite{goldfeld2020information}. There are only two notable exceptions. First, both $X$ and $Y$ have discrete alphabets, in which $T$ can be obtained with a generalized Blahut-Arimoto algorithm~\cite{blahut1972computation,arimoto1972algorithm}. Second, $X$ and $Y$ are jointly Gaussian, in which the solution reduces to a canonical correlation analysis (CCA) projection with tunable rank~\cite{chechik2005information}. 



The gap between the IB principle and its practical deep learning applications is mainly the result of the challenge in computing mutual information~\cite{goldfeld2020information,zaidi2020information,alemi2016deep}, a notoriously hard problem in high-dimensional space. 
Variational inference offers a natural solution to bridge the gap, as it constructs a lower bound on the IB objective which is tractable with the reparameterization trick~\cite{Kingma2014}. Notable examples in this direction include the deep variational information bottleneck (VIB)~\cite{alemi2016deep} and the $\beta$-variational autoencoder ($\beta$-VAE)~\cite{higgins2017beta}.



In this work, we provide a new neural network parameterization of IB principle. By making use of the recently proposed matrix-based R{\'e}nyi's $\alpha$-order entropy functional~\cite{giraldo2014measures,yu2019multivariate}, we show that one is able to explicitly train a deep neural network (DNN) by IB principle without variational approximation and distributional estimation. We term our methodology Deep Deterministic Information Bottleneck (DIB) and make the following contributions:

\begin{itemize}
    \item  We show that the matrix-based R{\'e}nyi's $\alpha$-order entropy functional is differentiable. This property complements the theory of this new family of estimators and opens the door for its deep learning applications.
    \item As a concrete example to demonstrate the advantage of this estimator, we apply it on representation learning and 
    show that it enables us to parameterize the IB principle with a deterministic neural network.
    \item We observed that the representation learned by DIB enjoys reduced generalization error and is more robust to adversarial attack than its variational counterpart.
\end{itemize}


\section{Related work}
\label{sec:related_work}

\subsection{IB Principle and Deep Neural Networks}



The application of IB principle on machine learning dates back to two decades ago, e.g., document clustering~\cite{slonim2000agglomerative} and image segmentation~\cite{bardera2009image}. 


Recently, the IB principle has been proposed for analyzing and understanding the dynamics of learning and the generalization of DNNs~\cite{shwartz2017opening}. \cite{elad2019direct} applies the recently proposed mutual information neural estimator (MINE)~\cite{belghazi2018mutual} to train hidden layer with IB loss, and freeze it before moving on to the next layer. Although the result corroborates partially the IB hypothesis in DNNs, some claims are still controversial~\cite{saxe2019information,yu2020understanding}.  
From a practical perspective, the IB principle has been used as a design tool for DNN classifiers and generative models. The VIB~\cite{alemi2016deep} parameterizes IB Lagrangian with a DNN via a variational lower bound and reparameterization trick. 
The nonlinear IB~\cite{kolchinsky2019nonlinear} uses a variational lower bound for $I(Y;T)$ and a non-parametric upper bound for $I(X;T)$. 
Information dropout~\cite{achille2018information} further argues that the IB principle promotes minimality, sufficiency and disentanglement of representations. On the other hand, the $\beta$-variational autoencoder ($\beta$-VAE)~\cite{higgins2017beta} is also formulated under an IB framework. It also enjoys a rate-distortion interpretation~\cite{burgess2018understanding}. 


\subsection{Matrix-based Entropy Functional and its Gradient}

We recap briefly the recently proposed matrix-based R{\'e}nyi's $\alpha$-order entropy functional on positive definite matrices. We refer interested readers to~\cite{giraldo2014measures,yu2019multivariate} for more details.

\begin{definition}
Let $\kappa:\chi \times \chi \mapsto \mathbb{R}$ be a real valued positive definite kernel that is also infinitely divisible~\cite{bhatia2006infinitely}. Given $\{\mathbf{x}_{i}\}_{i=1}^{n}\in \chi$, each $\mathbf{x}_i$ can be a real-valued scalar or vector, and the Gram matrix $K\in \mathbb{R}^{n\times n}$ computed as $K_{ij}=\kappa(\mathbf{x}_{i}, \mathbf{x}_{j})$, a matrix-based analogue to R{\'e}nyi's $\alpha$-entropy can be given by the following functional:
\begin{equation}\label{Renyi_entropy}
H_{\alpha}(A)=\frac{1}{1-\alpha}\log_2 \left(\tr (A^{\alpha})\right)=\frac{1}{1-\alpha}\log_{2}\left(\sum_{i=1}^{n}\lambda _{i}(A)^{\alpha}\right),
\end{equation}
where $\alpha\in (0,1)\cup(1,\infty)$. $A$ is the normalized version of $K$, i.e., $A=K/\tr (K)$. $\lambda _{i}(A)$ denotes the $i$-th eigenvalue of $A$.
\end{definition}


\begin{definition}
Given $n$ pairs of samples $\{\mathbf{x}_{i}, \mathbf{y}_{i}\}_{i=1}^{n}$, each sample contains two measurements $\mathbf{x}\in\chi$ and $\mathbf{y}\in\gamma$ obtained from the same realization. Given positive definite kernels $\kappa_{1}:\chi\times\chi\mapsto\mathbb{R}$ and $\kappa_{2}:\gamma\times\gamma\mapsto\mathbb{R}$, a matrix-based analogue to R{\'e}nyi's $\alpha$-order joint-entropy can be defined as:
\begin{equation}\label{Renyi_joint_entropy}
H_{\alpha}(A,B)=H_{\alpha}\left(\frac{A \circ B}{\tr (A \circ B)}\right),
\end{equation}
where $A_{ij}=\kappa_{1}(\mathbf{x}_{i}, \mathbf{x}_{j})$ , $B_{ij}=\kappa_{2}(\mathbf{y}_{i}, \mathbf{y}_{j})$ and $A\circ B$  denotes the Hadamard product between the matrices $A$ and $B$.
\end{definition}

Given Eqs.~(\ref{Renyi_entropy}) and (\ref{Renyi_joint_entropy}), the matrix-based R{\'e}nyi's $\alpha$-order mutual information $I_{\alpha}(A; B)$ in analogy of Shannon's mutual information is given by:
\begin{equation}\label{Renyi_MI}
I_{\alpha}(A;B)=H_{\alpha}(A)+H_{\alpha}(B)-H_{\alpha}(A,B).
\end{equation}

Throughout this work, we use the radial basis function (RBF) kernel $\kappa(\mathbf{x}_{i},\mathbf{x}_{j})=\exp(-\frac{\|\mathbf{x}_{i}-\mathbf{x}_{j}\|^{2}}{2\sigma ^{2}})$ to obtain the Gram matrices. 
For each sample, we evaluate its $k$ ($k=10$) nearest distances and take the mean. We choose kernel width $\sigma$ as the average of mean values for all samples.


As can be seen, the new family of estimators avoids the explicit estimation of the underlying distributions of data, which suggests its use in challenging problems involving high-dimensional data. Unfortunately, despite a few recent successful applications in feature selection~\cite{yu2019multivariate} and change detection~\cite{lv2020mutual}, its deep learning applications remain limited. One major reason is that its differentiable property is still unclear and under-investigated, which impedes its practical deployment as a loss function to train neural networks.

To bridge the gap, we first show that both $H_\alpha(A)$ and $H_\alpha(A,B)$ have analytical gradient. In fact, we have:
\begin{equation}
    \frac{\partial H_\alpha(A)}{\partial A} = \frac{\alpha}{(1-\alpha)} \frac{A^{\alpha-1}}{\mathrm{tr}\left(A^\alpha\right)},
\end{equation}
,
\begin{equation}
    \frac{\partial H_\alpha(A,B)}{\partial A} = \frac{\alpha}{(1-\alpha)}
    \left[ \frac{(A\circ B)^{\alpha-1}\circ B}{\mathrm{tr}(A\circ B)^{\alpha}}
    - \frac{I\circ B}{\mathrm{tr}(A\circ B)}
    \right]
\end{equation}
and
\begin{equation}
    \frac{\partial I_\alpha(A;B)}{\partial A} = \frac{\partial H_\alpha(A)}{\partial A} + \frac{\partial H_\alpha(A,B)}{\partial A}
\end{equation}
Since $I_\alpha(A;B)$ is symmetric, the same applies for $\frac{\partial I_\alpha(A;B)}{\partial B} $ with exchanged roles between $A$ and $B$.

In practice, taking the gradient of $I_{\alpha}(A;B)$ is simple with any automatic differentiation software, like PyTorch~\cite{paszke2019pytorch} or Tensorflow~\cite{abadi2016tensorflow}. We recommend PyTorch, because the obtained gradient is consistent with the analytical one.

\section{Deterministic Information Bottleneck}
\label{sec:methodology}

The IB objective contains two mutual information terms: $I(X;T)$ and $I(Y;T)$. When parameterizing IB objective with a DNN, $T$ refers to the latent representation of one hidden layer. In this work, we simply estimate $I(X;T)$ (in a mini-batch) with the above mentioned matrix-based R{\'e}nyi's $\alpha$-order entropy functional with Eq.~(\ref{Renyi_MI}).

The estimation of $I(Y;T)$ is different here. Note that $I(Y;T)=H(Y)-H(Y|T)$, in which $H(Y|T)$ is the conditional entropy of $Y$ given $T$. Therefore,
\begin{equation}
    \maximize I(T;Y)\Leftrightarrow \minimize H(Y|T).
\end{equation}
This is just because $H(Y)$ is a constant that is irrelevant to network parameters.

Let $p(\mathbf{x},\mathbf{y})$ denote the distribution of the training data, from which the training set $\{\mathbf{x}_{i}, \mathbf{y}_{i}\}_{i=1}^{N}$ is sampled. Also let $p_\theta(\mathbf{t}|\mathbf{x})$ and $p_\theta(\mathbf{y}|\mathbf{t})$ denote the unknown distributions that we wish to estimate, parameterized by $\theta$. We have~\cite{achille2018information}:
\begin{equation}
    H(Y|T)\simeq \mathbb{E}_{\mathbf{x},\mathbf{y}\sim p(\mathbf{x},\mathbf{y})}\left[\mathbb{E}_{t\sim p_\theta(\mathbf{t}|\mathbf{x})}\left[-\log p_\theta(\mathbf{y}|\mathbf{t})\right]\right].
\end{equation}

We can therefore empirically approximate it by:
\begin{equation}
    \frac{1}{N}\sum_{i=1}^N \mathbb{E}_{\mathbf{t}\sim p(\mathbf{t}|\mathbf{x}_i)}\left[-\log p(\mathbf{y}_i|\mathbf{t})\right],
\end{equation}
which is exactly the average cross-entropy loss~\cite{amjad2019learning}.

In this sense, our objective can be interpreted as a classic cross-entropy loss\footnote{The same trick has also been used in nonlinear IB~\cite{kolchinsky2019nonlinear} and VIB~\cite{alemi2016deep}.} regularized by a weighted differentiable mutual information term $I(X;T)$. 
We term this methodology Deep Deterministic Information Bottleneck (DIB)\footnote{We are aware of a previous work~\cite{strouse2017deterministic} that uses the same name of “deterministic information bottleneck” by maximizing the objective of $I(Y;T)-\beta H(T)$, where $H(T)$ is the entropy of $T$, an upper bound of the original $I(X;T)$. This objective discourages the the stochasticity in the mapping $p(t|x)$, hence seeking a ``deterministic" encoder function. We want to clarify here that the objective, the optimization and application domain in this work are totally different from~\cite{strouse2017deterministic}.}.

%


\section{Experiments}
\label{sec:experiments}


In this section, we perform experiments to demonstrate that: 1) models trained by our DIB objective converge well to the theoretical IB curve; 2) our DIB objective improves model generation performance and robustness to adversarial attack, compared to VIB and other forms of regularizations. 

\subsection{Information Bottleneck (IB) Curve}
Given two random variables $X$ and $Y$, and a “bottleneck” variable $T$. IB obeys the Markov condition that $I(X; T) \geq I(Y;T)$ based on the data processing inequality (DPI)~\cite{cover1999elements}, meaning that the bottleneck variable cannot contain more information about $Y$ than it does about $X$. 

\begin{figure}[htbp]
\centering
\subfigure[] {
    \includegraphics[width=0.23\textwidth]{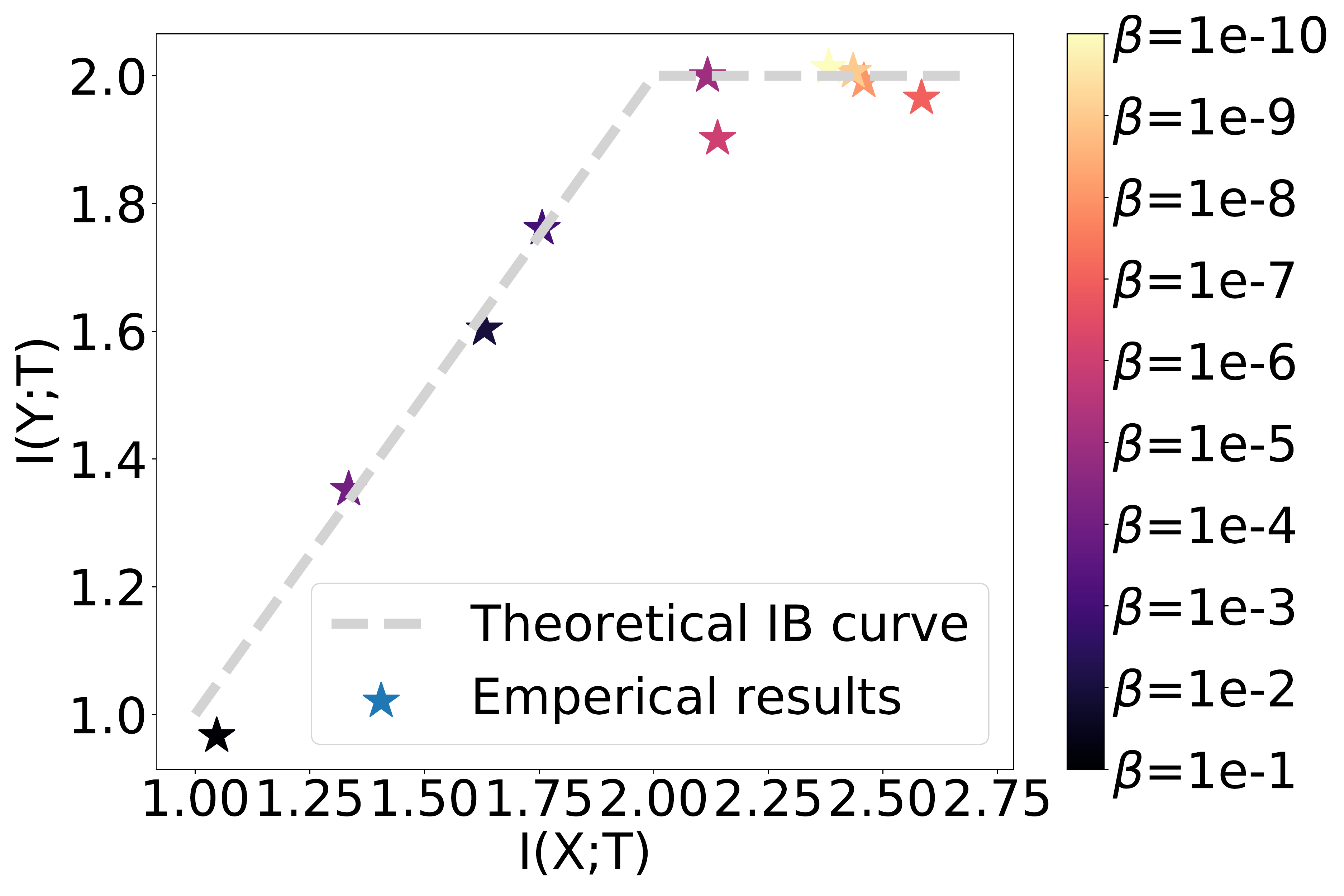}}
\subfigure[] {
    \includegraphics[width=0.23\textwidth]{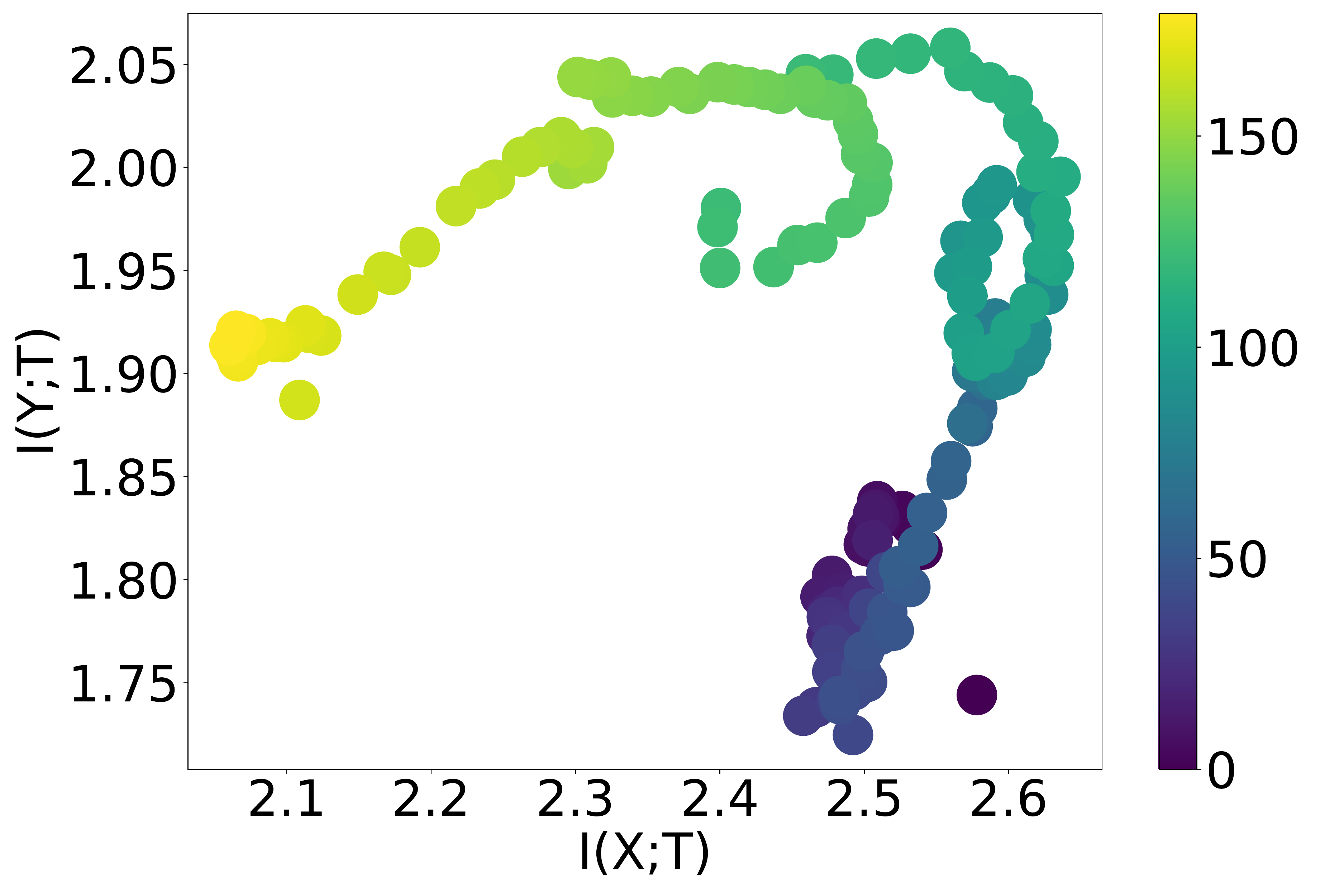}}
	\caption {(a) Theoretical (the dashed lightgrey line) and empirical IB curve found by maximizing the IB Lagrangian with different values of $\beta$; (b) a representative information plane for $\beta$=1$e$-6, different colors denote different training epochs.}
	\label{fig:fig4-1}
\end{figure}

According to~\cite{kolchinsky2018caveats}, the IB curve in classification scenario is piecewise linear and becomes a flat line at $I(Y;T) = H(Y)$ for $I(X;T) \geq H(Y)$.
We obtain both theoretical and empirical IB curve by training a three layer MLP with $256$ units in the bottleneck layer on MNIST dataset, as shown in  Fig.~\ref{fig:fig4-1}(a). As we can see, when $\beta$ is approaching to $0$, we place no constraint on the \textbf{minimality}, a representation learned in this case is sufficient for desired tasks but contains too much redundancy and nuisance factors. However, if $\beta$ is too large, we are at the risk of sacrificing the performance or representation \textbf{sufficiency}. Note that the region below the curve is feasible: for suboptimal mapping $p(t|x)$, solutions will lie in this region. No solution will lie above the curve.

We also plot a representative information plane~\cite{shwartz2017opening} (i.e., the values of $I(X;T)$ with respect to $I(Y;T)$ across the whole training epochs) with $\beta$=1$e$-6 in Fig.~\ref{fig:fig4-1}(b). It is very easy to observe the mutual information increase ($a.k.a.$, fitting) phase, followed by the mutual information decrease ($a.k.a.$, compression) phase. This result supports the IB hypothesis in DNNs.


\subsection{Image Classification}
\label{sec:4-2}
\subsubsection{MNIST}
\label{sec:4-2-1}
As a preliminary experiment, we first evaluate the performance of DIB objective on the standard MNIST digit recognition task. We randomly select $10k$ images from the training set as the validation set for hyper-parameter tuning.
For a fair comparison, we use the same architecture as has been adopted in~\cite{alemi2016deep}, namely a MLP with fully connected layers of the form $784-1024-1024-256-10$, and ReLU activation. The bottleneck layer is the one before the softmax layer, i.e., the hidden layer with $256$ units.
The Adam optimizer is used with an initial learning rate of 1$e$-4 and exponential decay by a factor of $0.97$ every $2$ epochs. All models are trained with $200$ epochs with mini-batch of size $100$. Table~\ref{table:Table1} shows the test error of different methods. DIB performs the best.


\vspace{-0.3cm}

 \setlength{\tabcolsep}{12pt}
\begin{table}[ht]

\centering
\fontsize{10}{12}\selectfont
\caption{Test error (\%) for permutation-invariant MNIST}
 \begin{threeparttable}[t]
\begin{tabular}[t]{lc}
\hline
\bf{Model}&\bf{Test} (\%)\\
\hline
Baseline&1.38\\
Dropout &1.28\\
Label Smoothing~\cite{pereyra2017regularizing}&1.24\\
Confidence Penalty~\cite{pereyra2017regularizing}&1.23\\
VIB~\cite{alemi2016deep}&1.17\tnote{1}\\
\midrule
{\bf DIB} ($\beta$=1$e$-6)&\bf{1.13}\\
\hline
\end{tabular}
 \begin{tablenotes}
     \item[1] \footnotesize{Result obtained on our test environment with authors’ original code \url{https://github.com/alexalemi/vib\_demo}.}
   \end{tablenotes}
    \end{threeparttable}%
\label{table:Table1}
\end{table}%

\vspace{-0.3cm}

\subsubsection{CIFAR-10}
CIFAR-10 is an image classification dataset consisting of $32\times32\times3$ RGB images of $10$ classes. As a common practice, We use $10k$ images in the training set for hyper-parameter tuning.
In our experiment, we use VGG16~\cite{simonyan2014very} as the baseline network and compare the performance of VGG16 trained by DIB objective and other regularizations. Again, we view the last fully connected layer before the softmax layer as the bottleneck layer. All models are trained
with $400$ epochs, a batch-size of $100$, and an initial learning rate $0.1$. The learning rate was reduced by a factor of $10$ for every $100$ epochs. We use SGD optimizer with weight decay
5$e$-4. We explored $\beta$ ranging from 1$e$-4 to $1$, and found that $0.01$ works the best. Test error rates with different methods are shown in Table~\ref{table:Table2}.
As can be seen, VGG16 trained with our DIB outperforms other regularizations and also the baseline ResNet50. We also observed, surprisingly, that VIB does not provide performance gain in this example, even though we use the authors' recommended value of $\beta$ ($0.01$). 

\vspace{-0.6cm}

\begin{table}[ht]

\centering
\fontsize{10}{12}\selectfont
\caption{Test error (\%) on CIFAR-10}
\begin{tabular}[t]{lc}
\hline
\bf{Model}&\bf{Test}(\%)\\
\hline
VGG16&7.36\\
ResNet18&6.98\\
ResNet50&6.36\\
\midrule
VGG16+Confidence Penalty
&5.75\\
VGG16+Label smoothing&5.78\\
VGG16+VIB&9.31\\
\midrule
{\bf VGG16+DIB} ($\beta$=1$e$-2)&\bf{5.66}\\
\hline
\end{tabular}
\label{table:Table2}
\end{table}%

\begin{figure}[htbp]
	\setlength{\abovecaptionskip}{0pt}
	\setlength{\belowcaptionskip}{0pt}
	\centering
	
	\subfigure[]{
		\includegraphics[width=4cm]{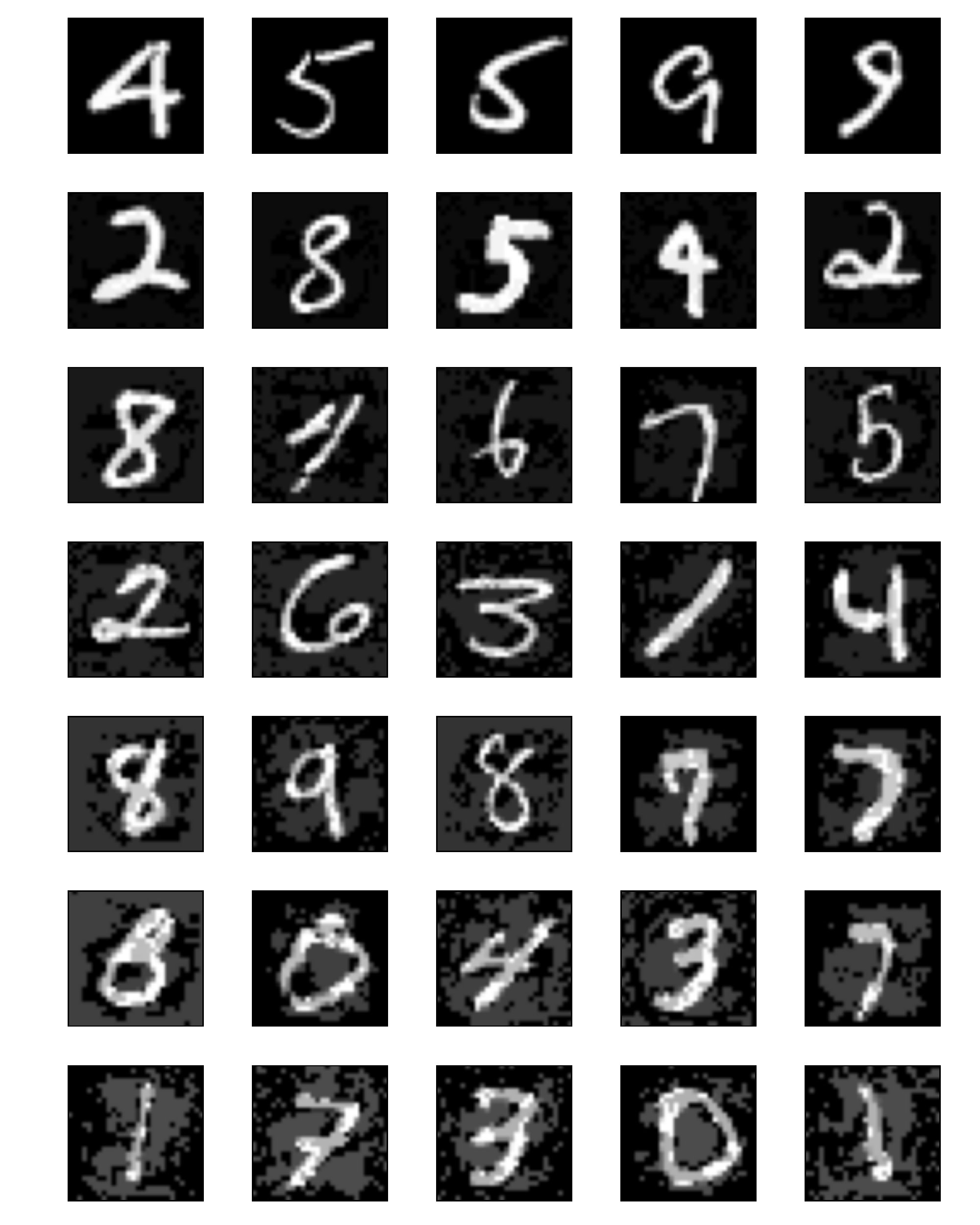}}
	\subfigure[]{
		\includegraphics[width=4cm]{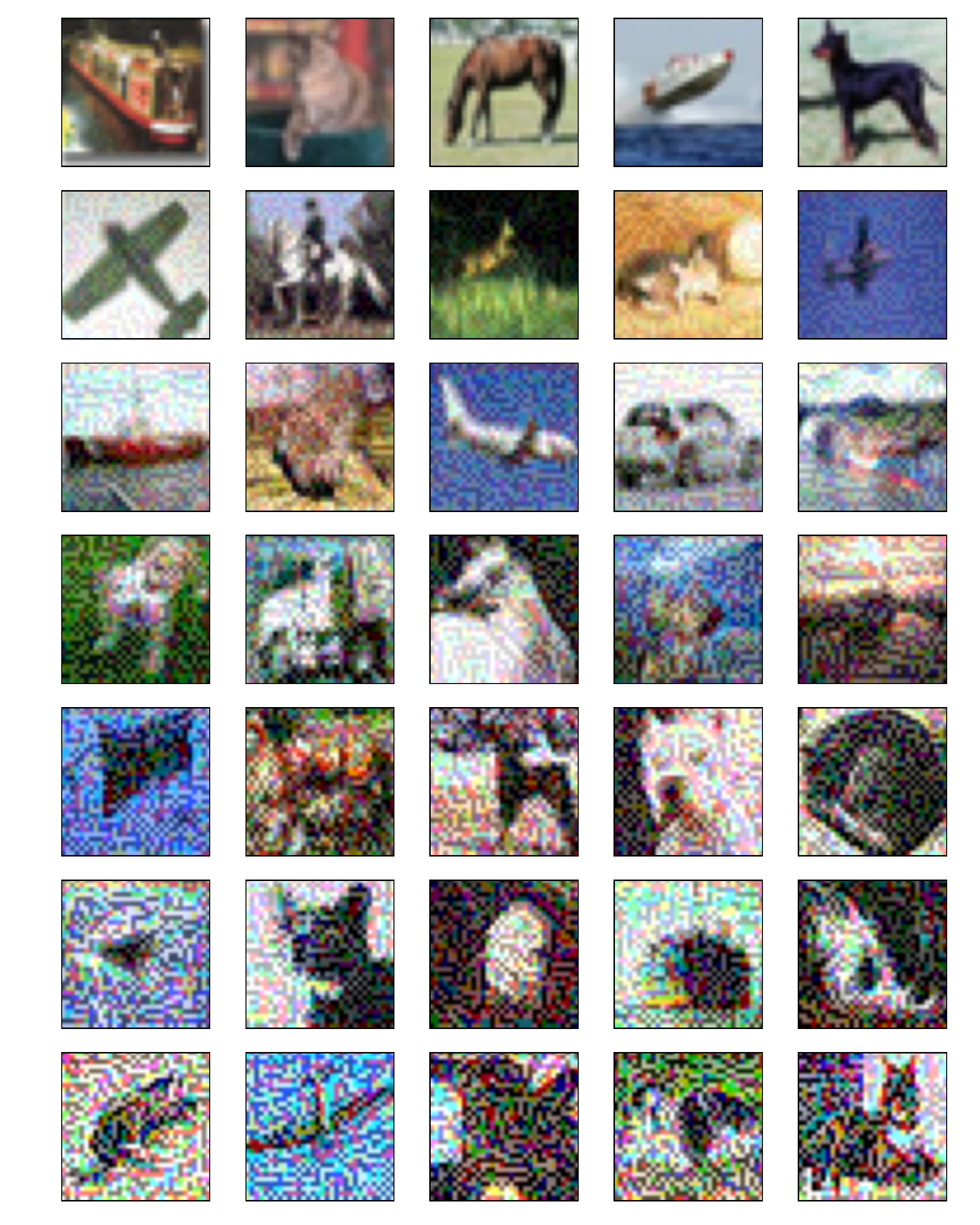}}
	\caption {Adversarial examples from (a) MNIST; (b) CIFAR-10 with different $\epsilon=[0,0.05,0.1,0.15,0.2,0.25,0.3]$ (from the first row to the last row).}
	\label{fig:fig4-3-1}
\end{figure}

\vspace{-0.5cm}

\subsection{Behavior on Adversarial Examples}

Our last experiment is to evaluate the adversarial robustness of model trained with our DIB objective. There are multiple definitions of adversarial robustness in the literature. The most basic one, which we shall use, is accuracy on adversarially perturbed versions of the test set, also called the adversarial examples. One of the most popular attack methods is the Fast Gradient Sign Attack (FGSM)~\cite{goodfellow2014explaining}, which uses the gradient of the objective function respect to the input image to generate an adversarial image maximizing the loss. The FGSM can be summarized by Eq.~(\ref{eq4-3}): 
\begin{equation}
\begin{split}
\hat{x} =  x + \epsilon \cdot sign(\nabla_{x} J(\mathbf{\theta}, \mathbf{x}, y)),
\end{split}
\label{eq4-3}
\end{equation}
where $x$ denotes the original clean image, $\epsilon$ is the pixel-wise perturbation amount, $\nabla_{x} J(\mathbf{\theta,\mathbf{x}, y)}$ is gradient of the loss with respect to the input image $x$, and $\hat{x}$ represents the perturbed image. Fig.~\ref{fig:fig4-3-1} shows some adversarial examples with different $\epsilon$ on MNIST and CIFAR-10 datasets.

\begin{figure}[htbp]
	\setlength{\abovecaptionskip}{0pt}
	\setlength{\belowcaptionskip}{0pt}
	\centering
	
	\subfigure[]{
		\includegraphics[width=4cm]{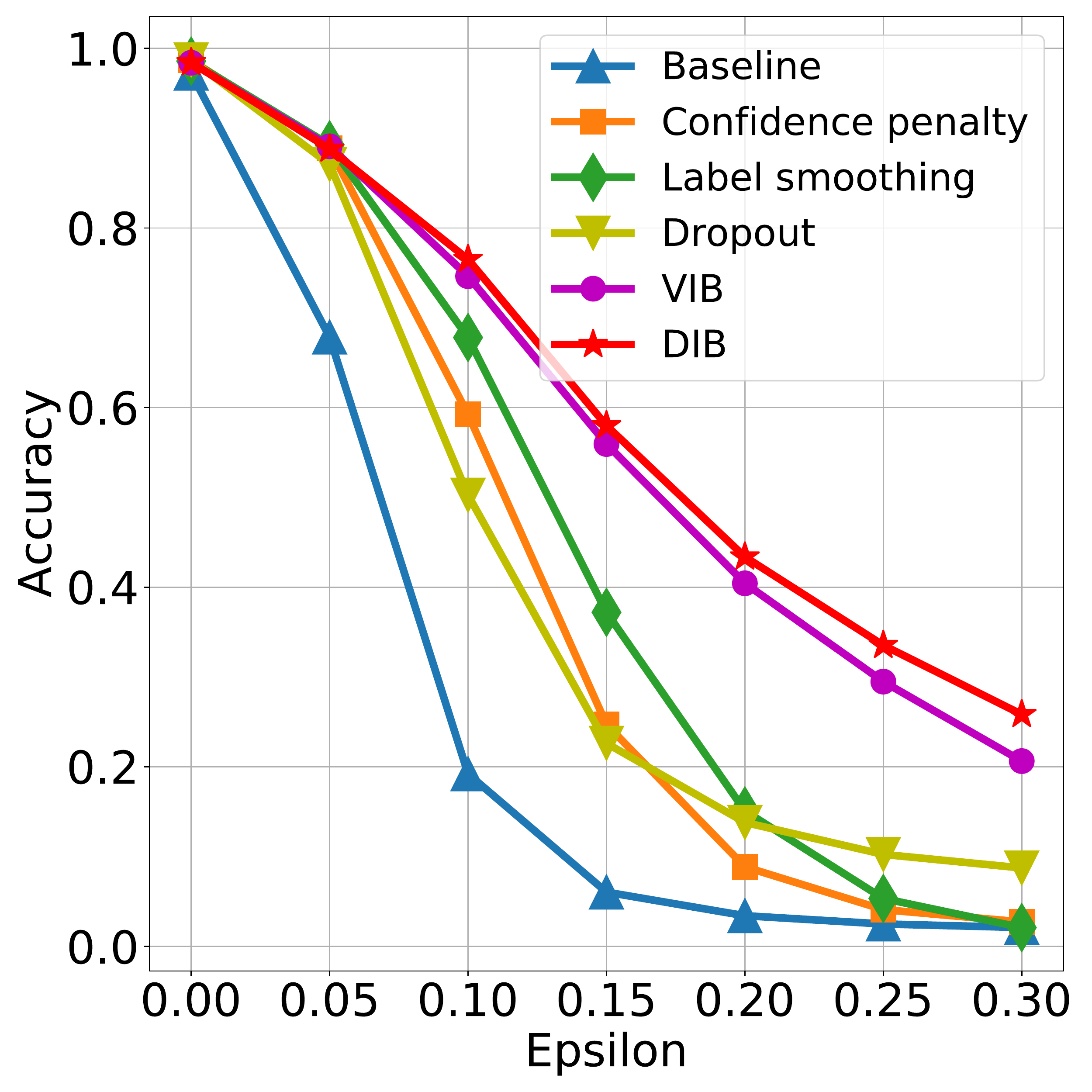}}
	\subfigure[]{
		\includegraphics[width=4cm]{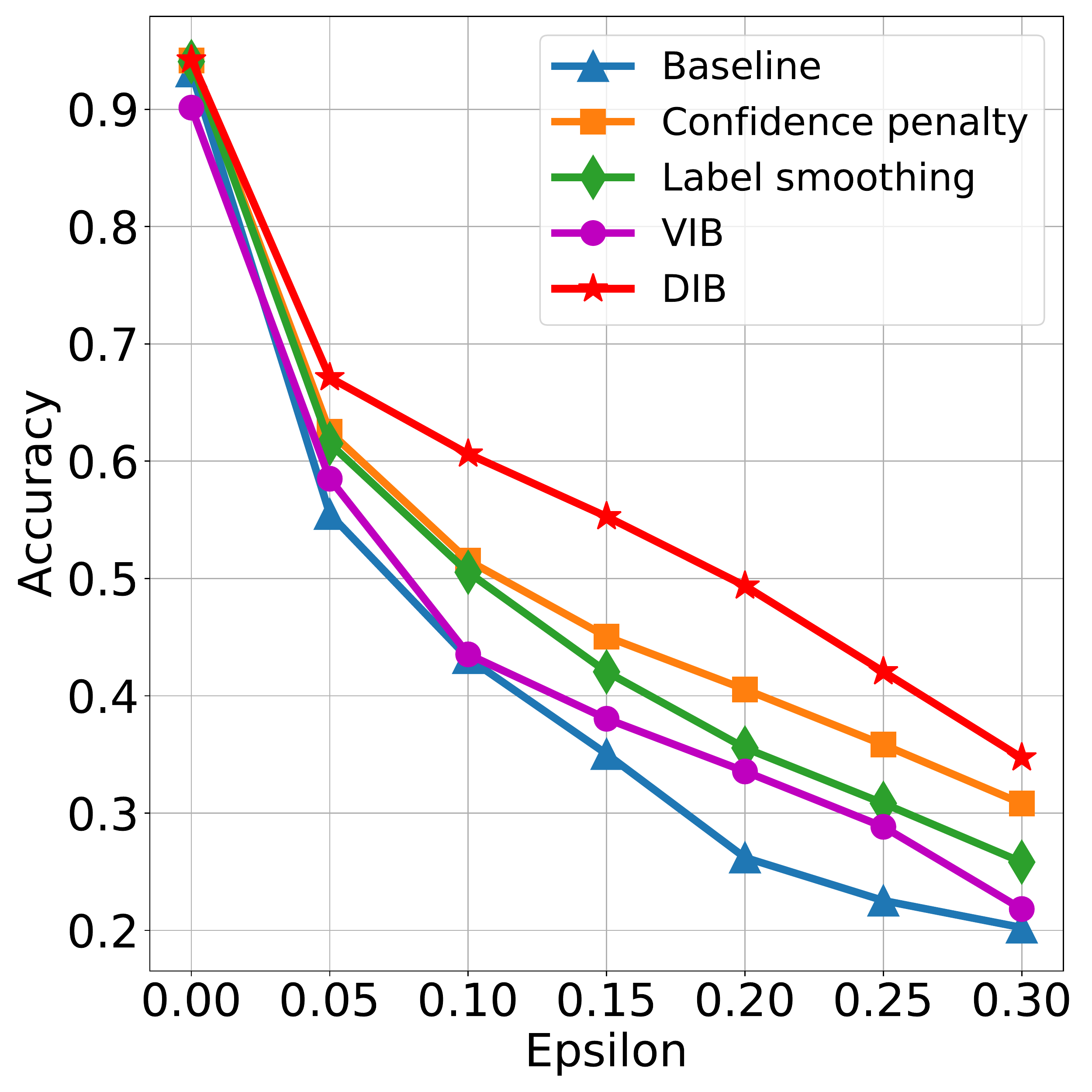}}
	\caption {Test accuracy with different $\epsilon$ and different methods on (a) MNIST; (b) CIFAR-10.}
	\label{fig:fig4-3-2}
\end{figure}

We compare behaviors of model trained with different forms of regularizations on MNIST and CIFAR-10 datasets under FGSM. We use the same experimental set up as in Section \ref{sec:4-2}, and only add adversarial attack on the test set. The results are shown in Fig.~\ref{fig:fig4-3-2}. As we can see, our DIB performs slightly better than VIB in MNIST, but is much better in CIFAR-10. In both datasets, our DIB is much more robust than label smoothing and confidence penalty.


\vspace{-0.5cm}

\section{Conclusions and Future Work}

We applied the matrix-based R{\'e}nyi's $\alpha$-order entropy functional to parameterize the IB principle with a neural network. The resulting DIB improved model's generalization and robustness.
In the future, we are interested in training a DNN in a layer-by-layer manner by the DIB objective proposed in this work. The training moves sequentially from lower layers to deeper layers. Such training strategy has the potential to avoid backpropagation and its related issues like the gradient vanishing. Meanwhile, we also plan to integrate the new family of estimators with the famed InfoMax principle~\cite{linsker1988self} to learn informative (and possibly disentangled) representations in a fully unsupervised manner. The same proposal has been implemented in Deep InfoMax (DIM)~\cite{hjelm2018learning} with MINE. However, we expect a performance gain due to the simplicity of the new estimator.


\bibliographystyle{IEEEbib}
{\fontsize{9}{10}\selectfont \bibliography{strings,refs}}

\end{document}